# Benchmark Problems for CEC2021 Competition on Evolutionary Transfer Multiobjectve Optimization

Songbai Liu[1], Qiuzhen Lin[2], Kay Chen Tan[3], Qing Li[3]
[1]Department of Computer Science, City University of Hong Kong, Hong Kong SAR
[2]College of Computer Science and Software Engineering, Shenzhen University, Shenzhen, China
[3]Department of Computing, The Hong Kong Polytechnic University, Hong Kong SAR

**Technical Report**

**Abstract**-Evolutionary transfer multiobjective optimization (ETMO) has been becoming a hot research topic in the field of evolutionary computation, which is based on the fact that knowledge learning and transfer across the related optimization exercises can improve the efficiency of others. Besides, the potential for transfer optimization is deemed invaluable from the standpoint of human-like problem-solving capabilities where knowledge gather and reuse are instinctive. To promote the research on ETMO, benchmark problems are of great importance to ETMO algorithm analysis, which helps designers or practitioners to understand the merit and demerit better of ETMO algorithms. Therefore, a total number of 40 benchmark functions are proposed in this report, covering diverse types and properties in the case of knowledge transfer, such as various formulation models, various PS geometries and PF shapes, large-scale of variables, dynamically changed environment, and so on. All the benchmark functions have been implemented in JAVA code, which can be downloaded in the following website: https://github.com/songbai-liu/etmo.

**Index Terms**-Evolutionary Transfer Optimization, Multiobjective Optimization, Multitasking, Benchmarks.

## 1. Introduction

Evolutionary algorithms (EAs) characterized with a population-based iterative search engine have been recognized as an effective tool for solving many complex multiobjective optimization problems (MOPs) [1]-[2], such as community detection, cyber security, feature selection in classification, and compressing deep neural networks. Despite the great success enjoyed by EAs, present EA-based solvers generally start their search from a completely random population, which means that the search start from scratch or the ground-zero state when given a new MOP, regardless how similar it is to the already addressed MOPs [3]. However, many real-world MOPs are closely related, so useful experience or knowledge are capable of learning from well-solved related MOPs, aiming to guide the search effectively for solving the new MOP [4]. In that regard, existing EAs have remained yet to exploit the useful knowledge fully that may exist in similar MOPs. Thus, a significantly under-explored area of evolutionary multiobjective optimization is the study of EA-based methodologies that can evolve along with the MOPs solved, i.e., the apparent lack of automated knowledge transfers and reuse across different MOPs with certain similarities [5]. Exactly, humans have an inherent ability to transfer knowledge across tasks/problems. What we acquire as knowledge when learning about one problem can be qualified in the same way to solve other related problems. The more related the problems are, the easier it is for us to transfer or cross-utilize our knowledge. In general, experience is "the best teacher" for the evolutionary search [6].

   Inspired by transfer learning which can reuse past experiences to solve relevant problems, transfer learning-based methods are widely used in evolutionary optimization [7], and evolutionary transfer optimization (ETO) has become a new frontier in evolutionary computation research [8]. Generally, when applying transfer-learning methods to solve MOPs, three key points should be considered [9]: 1) Transferability, i.e., the ability to avoid negative transfer. 2) Transfer components, i.e., identify which potential knowledge from the related MOPs is transferable and useful. 3)

Transfer method, i.e., reusing the learned knowledge effectively to help optimize the new MOP. Thoughtfully, ETO is defined as a paradigm that integrates EA algorithms with knowledge learning and transfer across related domains [10] as shown in Fig. 1, aiming to improve the optimization performance when solving a variety of problems, such as multi-/many-objective optimization problems (MOPs), dynamic optimization problems (DOPs), multi-task optimization problems (MTOPs), large-scale optimization problems (LMOPs), expensive optimization problems (EOPs), and other real-world complex optimization problems (COPs), etc. Intuitively, ETO provides a shortcut to solve the target problem via reusing the experiences learned from the optimization exercises of solving the source problems, which can bring the following three possible benefits:

- **Head start**: get an initial population with better performance and robustness without starting from scratch search.
- **Faster convergence speed**: solve the target problem with faster convergence speed and fewer computational costs.
- **Closer approximation:** get a final population with better approximation to the optima for the target problem.

To promote the research on evolutionary transfer multiobjective optimization (ETMO), benchmark problems are of great importance to ETMO algorithm analysis, which helps designers or practitioners to understand the merit and demerit better of ETMO algorithms. However, although there are many areas that ETMO can cover, there are few types of benchmark problems that exist. In this report, a total of 40 benchmark functions are introduced, covering diverse types and properties in the case of knowledge transfer, such as various formulation models, various PS geometries and PF shapes, large-scale of variables, dynamically changed environment, and so on. All the benchmark functions have been implemented in JAVA code based on the codes provided by [11], which can be downloaded in the following website: https://github.com/songbai-liu/etmo.

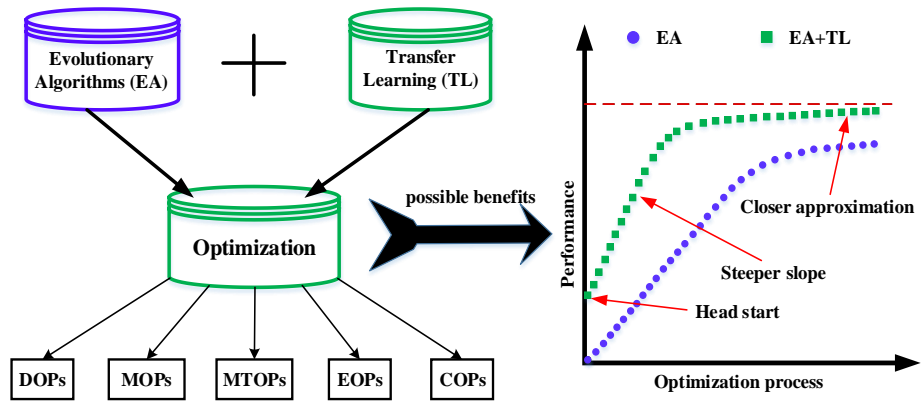

Fig. 1 Illustration of the paradigm for evolutionary transfer optimization.

## 2. Summary of 40 Test Problems

In this report, the proposed test suite (called ETMOF) has 40 benchmark problems for fully testing the performance of ETMO algorithms. In particular, these 40 benchmark problems can be classified as the following five types:

(1) Evolutionary Transfer Multiobjective Optimization Problems: ETMOF1 to ETMOF8

(2) Evolutionary Transfer Many-objective Optimization Problems: ETMOF9 to ETMOF16

(3) Evolutionary Transfer Large-scale Multiobjective Optimization Problems: ETMOF17 to ETMOF24

(4) Evolutionary Transfer Many-Task Optimization Problems: ETMOF25 to ETMOF32

(5) Evolutionary Transfer Dynamic Multiobjective Optimization Problems: ETMOF33 to ETMOF40

## 3. Definitions of Basic Components

In this report, a multiobjective optimization problem has $m$ objectives, i.e., $F(x) = (f_1(x), f_2(x), ..., f_m(x))$, where the solution vector x has $n$ variables, i.e., $x = (x_1, x_2, ..., x_n)$.

### 3.1 Definitions of various formulation models based on deep grouping of variables

**First layer grouping**: the $n$ variables are linearly divided into two variable groups, i.e., a group with $K$ position-related variables: $x^p = (x_1^p, x_2^p, ..., x_K^p)$ and another group with $L$ distance-related variables: $x^d = (x_1^d, x_2^d, ..., x_L^d)$, $n = K + L$. In this layer of variable groups, we can get the first multiplication-based formulation model used in this report to design the $m$ objective functions, as below:

$$F_1(x): \min \begin{cases} f_1(x) = h_1(x^p)(1 + g(x^d)) \\ f_2(x) = h_2(x^p)(1 + g(x^d)) \\ ... \\ f_m(x) = h_m(x^p)(1 + g(x^d)) \end{cases},$$

where functions $h_1$ to $h_m$ together define the shape of the PF, known as the shape functions, and function $g$ define the fitness landscape, known as the landscape function. Obviously, all of these $m$ objective functions in $F_1(x)$ are formulated by sharing the same landscape function that is associated with all of the $L$ distance-related variables.

**Second layer grouping** [19]: The $K$ position-related variables are further uniformly divided into $m-1$ variable groups: $x^{p,1}$ to $x^{p,m-1}$, where each group has its own unique variables. Besides, overlap variables may exist between neighbor groups and shared variables may exist in all groups. Overlap variables between two different groups in this case means that only in these two groups can these variables be found, and shared variables can be found in every group, while the unique variable of a group refers to the variable that can only be found in this group. Similarly, the $L$ distance-related variables are further unevenly divided into $m$ variable groups: $x^{d,1}$ to $x^{d,m}$. Based on this layer of variable groups, we can get four different formulation models used in this report to design the $m$ objective functions, as follows,

$$F_2(x): \min \begin{cases} f_1(x) = h_1(x^p)(1 + g_1(x^{d,1})) \\ f_2(x) = h_2(x^p)(1 + g_2(x^{d,2})) \\ ... \\ f_m(x) = h_m(x^p)(1 + g_m(x^{d,m})) \end{cases}, F_3(x): \min \begin{cases} f_1(x) = h_1(x^p) + g_1(x^{d,1}) \\ f_2(x) = h_2(x^p) + g_2(x^{d,2}) \\ ... \\ f_m(x) = h_m(x^p) + g_m(x^{d,m}) \end{cases}$$

$$F_4(x): \min \begin{cases} f_1(x) = h_1(x^p)(1 + \sum_{i=1}^m g_i(x^{d,i})) \\ f_2(x) = h_2(x^p)(1 + \sum_{i=1}^m g_i(x^{d,i})) \\ ... \\ f_m(x) = h_m(x^p)(1 + \sum_{i=1}^m g_i(x^{d,i})) \end{cases}, F_5(x): \min \begin{cases} f_1(x) = h_1(x^p) + \sum_{i=1}^m g_i(x^{d,i}) \\ f_2(x) = h_2(x^p) + \sum_{i=1}^m g_i(x^{d,i}) \\ ... \\ f_m(x) = h_m(x^p) + \sum_{i=1}^m g_i(x^{d,i}) \end{cases}$$

$F_2(x)$ and $F_4(x)$ are two multiplication-based formulation, $F_3(x)$ and $F_5(x)$ are two addition-based formulation.

**Third layer grouping**: variables in the $i$th distance-related two-layer variable group $x^{d,i}$ are further divided into $q_i$ subgroups: $x^{d,i,1}, x^{d,i,2}, ..., x^{d,i,q_i}$, $i = 1, 2, ..., m$. According to this layer of variable groups, we can get the last two formulation models used in this report for the $m$ objective functions, as below:

$$F_6(\mathbf{x}): \min \begin{cases} f_1(\mathbf{x}) = h_1(\mathbf{x}^p)(1+1/q_1 \sum_{j=1}^{q_1} g_{1,j}(\mathbf{x}^{d,1,j})) \\ f_2(\mathbf{x}) = h_2(\mathbf{x}^p)(1+1/q_2 \sum_{j=1}^{q_2} g_{2,j}(\mathbf{x}^{d,2,j})) \\ \dots \\ f_m(\mathbf{x}) = h_m(\mathbf{x}^p)(1+1/q_m \sum_{j=1}^{q_m} g_{m,j}(\mathbf{x}^{d,m,j})) \end{cases}, \quad F_7(\mathbf{x}): \min \begin{cases} f_1(\mathbf{x}) = h_1(\mathbf{x}^p)+1/q_1 \sum_{j=1}^{q_1} g_{1,j}(\mathbf{x}^{d,1,j}) \\ f_2(\mathbf{x}) = h_2(\mathbf{x}^p)+1/q_2 \sum_{j=1}^{q_2} g_{2,j}(\mathbf{x}^{d,2,j}) \\ \dots \\ f_m(\mathbf{x}) = h_m(\mathbf{x}^p)+1/q_m \sum_{j=1}^{q_m} g_{m,j}(\mathbf{x}^{d,m,j}) \end{cases}$$

$F_6(\mathbf{x})$ and $F_7(\mathbf{x})$ are mainly used to design the large-scale multiobjective optimization problems.

## 3.2 Definitions of the instances of Pareto Front (PF), i.e., the Shape Functions

Suppose $y_i = \left\| \frac{1}{|\mathbf{x}^{p,i}|} \sum_{j=1}^{|\mathbf{x}^{p,i}|} x_j^{p,i} \right\|$, $x_j^{p,i} \in [-1,1]$, $i=1,2,\dots,m-1$

1) Convex PF in the case of $m = 2$: $H_1(\mathbf{x}^p): \begin{cases} h_1(\mathbf{x}^p) = y_1 \\ h_2(\mathbf{x}^p) = 1-\sqrt{y_1} \end{cases}$, $y_1 \in [0,1]$

2) Nonconvex PF in the case of $m = 2$: $H_2(\mathbf{x}^p): \begin{cases} h_1(\mathbf{x}^p) = y_1 \\ h_2(\mathbf{x}^p) = 1-(y_1)^2 \end{cases}$, $y_1 \in [0,1]$

3) Linear PF:

$$H_3(\mathbf{x}^p): \begin{cases} h_1(\mathbf{x}^p) = y_1 y_2 \dots y_{m-1} \\ h_2(\mathbf{x}^p) = y_1 y_2 \dots (1-y_{m-1}) \\ \dots \\ h_{m-1}(\mathbf{x}^p) = y_1(1-y_2) \\ h_m(\mathbf{x}^p) = (1-y_1) \end{cases}, \quad y_i \in [0,1]$$

4) Inverted Linear PF:

$$H_4(\mathbf{x}^p): \begin{cases} h_1(\mathbf{x}^p) = 1 - y_1 y_2 \dots y_{m-1} \\ h_2(\mathbf{x}^p) = 1 - y_1 y_2 \dots (1-y_{m-1}) \\ \dots \\ h_{m-1}(\mathbf{x}^p) = 1 - y_1(1-y_2) \\ h_m(\mathbf{x}^p) = y_1 \end{cases}, \quad y_i \in [0,1]$$

5) Sphere PF:

$$H_5(\mathbf{x}^p): \begin{cases} h_1(\mathbf{x}^p) = \cos(0.5\pi y_1) \dots \cos(0.5\pi y_{m-2}) \cos(0.5\pi y_{m-1}) \\ h_2(\mathbf{x}^p) = \cos(0.5\pi y_1) \dots \cos(0.5\pi y_{m-2}) \sin(0.5\pi y_{m-1}) \\ \dots \\ h_{m-1}(\mathbf{x}^p) = \cos(0.5\pi y_1) \sin(0.5\pi y_2) \\ h_m(\mathbf{x}^p) = \sin(0.5\pi y_1) \end{cases}, \quad y_i \in [0,1]$$

6) Inverted Sphere PF:

$$H_6(\mathbf{x}^p): \begin{cases} h_1(\mathbf{x}^p) = 1 - \cos(0.5\pi y_1)...\cos(0.5\pi y_{m-2})\cos(0.5\pi y_{m-1}) \\ h_2(\mathbf{x}^p) = 1 - \cos(0.5\pi y_1)...\cos(0.5\pi y_{m-2})\sin(0.5\pi y_{m-1}) \\ ... \\ h_{m-1}(\mathbf{x}^p) = 1 - \cos(0.5\pi y_1)\sin(0.5\pi y_2) \\ h_m(\mathbf{x}^p) = 1 - \sin(0.5\pi y_1) \end{cases}, \; y_i \in [0,1]$$

7) Convex PF in the case of $m \geq 3$

$$H_7(\mathbf{x}^p): \begin{cases} h_1(\mathbf{x}^p) = [\cos(0.5\pi y_1)...\cos(0.5\pi y_{m-2})\cos(0.5\pi y_{m-1})]^4 \\ h_2(\mathbf{x}^p) = [\cos(0.5\pi y_1)...\cos(0.5\pi y_{m-2})\sin(0.5\pi y_{m-1})]^4 \\ ... \\ h_{m-1}(\mathbf{x}^p) = [\cos(0.5\pi y_1)\sin(0.5\pi y_2)]^4 \\ h_m(\mathbf{x}^p) = [\sin(0.5\pi y_1)]^2 \end{cases}, \; y_i \in [0,1]$$

8) Degenerated PF used in $F_1(\mathbf{x})$:

$$H_8(\mathbf{x}^p): \begin{cases} h_1(\mathbf{x}^p) = \cos(\theta_1)...\cos(\theta_{m-2})\cos(\theta_{m-1}) \\ h_2(\mathbf{x}^p) = \cos(\theta_1)...\cos(\theta_{m-2})\sin(\theta_{m-1}) \\ ... \\ h_{m-1}(\mathbf{x}^p) = \cos(\theta_1)\sin(\theta_2) \\ h_m(\mathbf{x}^p) = \sin(\theta_1) \end{cases}, \; \theta_i = \begin{cases} 0.5 y_i, & i = 1 \\ \dfrac{\pi}{4[1+g(\mathbf{x}^d)]}[1 + 2g(\mathbf{x}^d)y_i], & i > 1 \end{cases}$$

9) Irregular Concave PF:

$$H_9(\mathbf{x}^p): \begin{cases} h_1(\mathbf{x}^p) = \cos(\theta_1)...\cos(\theta_{m-2})\cos(\theta_{m-1}) \\ h_2(\mathbf{x}^p) = \cos(\theta_1)...\cos(\theta_{m-2})\sin(\theta_{m-1}) \\ ... \\ h_{m-1}(\mathbf{x}^p) = \cos(\theta_1)\sin(\theta_2) \\ h_m(\mathbf{x}^p) = \sin(\theta_1) \end{cases}, \; \theta_i = \dfrac{\pi}{2}(\dfrac{y_i}{2} + \dfrac{1}{4})$$

10) Disconnected PF used in $F_1(\mathbf{x})$:
$$H_{10}(\mathbf{x}^p): \begin{cases} h_1(\mathbf{x}^p) = \dfrac{y_1}{1+g(\mathbf{x}^d)} \\ h_2(\mathbf{x}^p) = \dfrac{y_2}{1+g(\mathbf{x}^d)} \\ ... \\ h_{m-1}(\mathbf{x}^p) = \dfrac{y_{m-1}}{1+g(\mathbf{x}^d)} \\ h_m(\mathbf{x}^p) = m - \sum_{i=1}^{m-1} \dfrac{y_i(1+\sin(3\pi y_i))}{1+g(\mathbf{x}^d)} \end{cases}$$

**3.3 Definitions of the Basic Single Objective Functions, i.e., the main components of the landscape**

**function**

1) Sphere Function: $b_1(x) = \sum_{i=1}^{|x|} (x_i)^2$

2) High Conditioned Elliptic Function: $b_2(x) = \sum_{i=1}^{|x|} [10^{\frac{i-1}{|x|-1}} (x_i)^2]$

3) Bent Cigar Function: $b_3(x) = (x_1)^2 + 10^6 \sum_{i=2}^{|x|} (x_i)^2$

4) Discus Function: $b_4(x) = 10^6 (x_1)^2 + \sum_{i=2}^{|x|} (x_i)^2$

5) Rosenbrock's Function: $b_5(x) = \sum_{i=1}^{|x|-1} \{100[(x_i)^2 - x_{i+1}]^2 + (x_i - 1)^2\}$

6) Ackley's Function

$$b_6(x) = 20 - 20\exp\left(-0.2\sqrt{\frac{1}{|x|}\sum_{i=1}^{|x|}(x_i)^2}\right) + e - \exp\left(\frac{1}{|x|}\sum_{i=1}^{|x|}(\cos 2\pi x_i)\right)$$

7) Weierstrass Function

$$b_7(x) = \sum_{i=1}^{|x|}\left(\sum_{k=0}^{20}\left[\alpha^k \cos\left(w\beta^k (x_i + 0.5)\right)\right]\right) - |x|\sum_{k=0}^{20}[\alpha^k \cos(0.5 w \beta^k)], \; \alpha=0.5, \beta=3, w=2\pi$$

8) Griewank's Function: $b_8(x) = \sum_{i=1}^{|x|} \frac{(x_i)^2}{4000} - \prod_{i=1}^{|x|} \cos(\frac{x_i}{\sqrt{i}}) + 1$

9) Rastrigin's Function: $b_9(x) = \sum_{i=1}^{|x|} [(x_i)^2 - 10\cos(2\pi x_i) + 10]$

The detail properties of these basic single objective functions can be found in [12].

10) Modified Schwefel's Function

$$b_{10}(x) = 418.9829|x| - \sum_{i=1}^{|x|} u(z_i) \qquad z_i = x_i + 4.209687462275036e+002$$

$$u(z_i) = \begin{cases} z_i \sin\left(|z_i|^{0.5}\right) & \text{if } \|z_i\| \leq 500 \\ [500 - \text{mod}(z_i, 500)]\sin\left(\sqrt{\|500 - \text{mod}(z_i, 500)\|}\right) - \frac{(z_i - 500)^2}{10000|x|} & \text{if } z_i > 500 \\ [\text{mod}(\|z_i\|, 500) - 500]\sin\left(\sqrt{\|\text{mod}(\|z_i\|, 500) - 500\|}\right) - \frac{(z_i + 500)^2}{10000|x|} & \text{if } z_i < -500 \end{cases}$$

11) Katsuura Function

$$b_{11}(\mathbf{x}) = \frac{10}{|\mathbf{x}|^2} \prod_{i=1}^{|\mathbf{x}|} \left( 1 + i \sum_{j=1}^{32} \frac{\|2^j x_i - round(2^j x_i)\|}{2^j} \right)^{\frac{10}{|\mathbf{x}|^{1.2}}} - \frac{10}{|\mathbf{x}|^2}$$

12) HappyCat Function

$$b_{12}(\mathbf{x}) = \left\| \sum_{i=1}^{|\mathbf{x}|} (x_i)^2 - |\mathbf{x}| \right\|^{1/4} + \frac{0.5 \sum_{i=1}^{|\mathbf{x}|} (x_i)^2 + \sum_{i=1}^{|\mathbf{x}|} x_i}{|\mathbf{x}|} + 0.5$$

13) Expanded Griewank's plus Rosenbrock's Function

$$b_{13}(\mathbf{x}) = b_8(b_5(x_1, x_2)) + b_8(b_5(x_2, x_3)) + \ldots + b_8(b_5(x_{|\mathbf{x}|-1}, x_{|\mathbf{x}|})) + b_8(b_5(x_{|\mathbf{x}|}, x_1))$$

14) Absolute Mean Function: $b_{14}(\mathbf{x}) = \frac{9}{|\mathbf{x}|} \sum_{i=1}^{|\mathbf{x}|} \|x_i\|$

15) Linkage Function 1

$$L_1(\mathbf{x}^d, y_1) = \frac{2}{|\mathbf{x}^d|} \sum_{i=1}^{|\mathbf{x}^d|} \left[ x_i^d - \left( 0.3(y_1)^2 \cos\left( 24\pi y_1 + \frac{4 \, index(x_i^d)\pi}{n} \right) + 0.6 y_1 \right) \left( \sin\left( 6\pi y_1 + \frac{index(x_i^d)\pi}{n} \right) \right) \right]^2$$

where $index(x_i^d)$ represents the original ordinal of variable $x_i^d$ in x hereafter

16) Linkage Function 2

$$L_2(\mathbf{x}^d, y_1) = \frac{2}{|\mathbf{x}^d|} \sum_{i=1}^{|\mathbf{x}^d|} \left[ x_i^d - (y_1)^{0.5\left(1.0 + \frac{3(index(x_i^d)-2)}{n-2}\right)} \right]^2, \quad index(x_i^d) \text{ represents the original ordinal of variable } x_i^d \text{ in x}$$

17) Linkage Function 3

$$L_3(\mathbf{x}^d, y_1) = \frac{2}{|\mathbf{x}^d|} \left[ 4 \sum_{i=1}^{|\mathbf{x}^d|} \left[ x_i^d - (y_1)^{0.5\left(1.0 + \frac{3(index(x_i^d)-2)}{n-2}\right)} \right]^2 - 2 \prod_{i=1}^{|\mathbf{x}^d|} \cos\left( \frac{20\pi \left( x_i^d - (y_1)^{0.5\left(1.0 + \frac{3(index(x_i^d)-2)}{n-2}\right)} \right)}{\sqrt{index(x_i^d)}} \right) + 2 \right]$$

18) Linkage Function 4

$$L_4(\mathbf{x}^d, y_1) = \frac{2}{|\mathbf{x}^d|} \sum_{i=1}^{|\mathbf{x}^d|} \left[ x_i^d - 0.8 y_1 \cos\left( 6\pi y_1 + \frac{index(x_i^d)\pi}{n} \right) \right]^2$$

19) Linkage Function 5

$$L_5(\mathbf{x}^d, y_1) = \frac{2}{|\mathbf{x}^d|} \sum_{i=1}^{|\mathbf{x}^d|} \left[ x_i^d - 0.8 y_1 \sin\left( 6\pi y_1 + \frac{index(x_i^d)\pi}{n} \right) \right]^2$$

20) Linkage Function 6

$$L_6(\mathbf{x}^d, y_1, y_2) = \frac{2}{|\mathbf{x}^d|} \sum_{i=1}^{|\mathbf{x}^d|} \left[ x_i^d - 2y_2 \sin\left(2\pi y_1 + \frac{index(x_i^d)\pi}{n}\right)\right]^2$$

21) Linkage Function 7

$$L_7(\mathbf{x}^d, y_1) = \frac{2}{|\mathbf{x}^d|} \sum_{i=1}^{|\mathbf{x}^d|} \left[ x_i^d - \sin\left(6\pi y_1 + \frac{index(x_i^d)\pi}{n}\right)\right]^2$$

22) Linkage Function 8

$$L_8(\mathbf{x}^d, y_1) = \frac{2}{|\mathbf{x}^d|} \sum_{i=1}^{|\mathbf{x}^d|} \left[ 4\left[x_i^d - (y_1)^{0.5\left(1.0+\frac{3(index(x_i^d)-2)}{n-2}\right)}\right]^2 - \cos\left[8\pi\left(x_i^d - (y_1)^{0.5\left(1.0+\frac{3(index(x_i^d)-2)}{n-2}\right)}\right)\right] + 1\right]$$

## 4. Definitions of 40 Test Instances

1) **ETMOF1** is a 2-task optimization problem, where task $T_1$ is a 2-objective convex problems and task $T_2$ is a 2-objective concave problems. $T_1$ and $T_2$ are defined as follows:

$$T_1 : \begin{cases} \text{Formulation model: } F_2(\mathbf{x}), m=2, n=50, K=1, L=49 \\ \text{ShapeFunction: } H_1(\mathbf{x}^p) \\ \text{LandscapeFunction: } g(\mathbf{x}^d) = L_1(rotation(\mathbf{x}^d), y_1), x_j^d \in [-10,10] \end{cases}$$

$$T_2 : \begin{cases} \text{Formulation model: } F_3(\mathbf{x}), m=2, n=50, K=1, L=49 \\ \text{ShapeFunction: } H_2(\mathbf{x}^p) \\ \text{LandscapeFunction: } g_i(\mathbf{x}^{d,i}) = L_1(rotation(\mathbf{x}^{d,i}), y_1), x_j^d \in [-10,10] \end{cases}$$

Different rotation matrixes are used in $T_1$ and $T_2$

2) **ETMOF2** is a 2-task optimization problem, where both task $T_1$ and task $T_2$ are 2-objective convex problems. $T_1$ and $T_2$ are defined as follows::

$$T_1 : \begin{cases} \text{Formulation model: } F_3(\mathbf{x}), m=2, n=50, K=1, L=49 \\ \text{ShapeFunction: } H_1(\mathbf{x}^p) \\ \text{LandscapeFunction: } g_i(\mathbf{x}^{d,i}) = L_2(rotation(\mathbf{x}^{d,i}), y_1), x_j^d \in [-10,10] \end{cases}$$

$$T_2 : \begin{cases} \text{Formulation model: } F_3(\mathbf{x}), m=2, n=50, K=1, L=49 \\ \text{ShapeFunction: } H_1(\mathbf{x}^p) \\ \text{LandscapeFunction: } g_i(\mathbf{x}^{d,i}) = L_3(rotation(\mathbf{x}^{d,i}), y_1), x_j^d \in [-10,10] \end{cases}$$

Different rotation matrixes are used in $T_1$ and $T_2$

3) **ETMOF3** is a 2-task optimization problem, where task $T_1$ is a 2-objective problem with a convex PF and task $T_2$ is a 3-objective optimization problem with a sphere PF. $T_1$ and $T_2$ are defined as follows:

$$T_1 : \begin{cases} \text{Formulation model: } F_3(x), m = 2, n = 50, K = 1, L = 49 \\ \text{ShapeFunction: } H_1(x^p) \\ \text{LandscapeFunction: } g_i(x^{d,i}) = \begin{cases} L_4(rotation(x^{d,i}), y_1), & \text{if } i \bmod 2 = 1 \\ L_5(rotation(x^{d,i}), y_1), & \text{otherwise} \end{cases}, x_j^d \in [-10,10] \end{cases}$$

$$T_2 : \begin{cases} \text{Formulation model: } F_3(x), m = 3, n = 51, K = 2, L = 49 \\ \text{ShapeFunction: } H_5(x^p) \\ \text{LandscapeFunction: } g_i(x^{d,i}) = L_6(rotation(x^{d,i}), y_1, y_2), x_j^d \in [-10,10] \end{cases}$$

Different rotation matrixes are used in $T_1$ and $T_2$

4) **ETMOF4** is a 2-task optimization problem, where both task $T_1$ and task $T_2$ are 3-objective problems with linear PF. $T_1$ and $T_2$ are defined as follows:

$$T_1 : \begin{cases} \text{Formulation model: } F_4(x), m = 3, n = 51, K = 2, L = 49 \\ \text{ShapeFunction: } H_3(x^p) \\ \text{LandscapeFunction: } g_i(x^{d,i}) = \begin{cases} b_3(shiftRotation(x^{d,i})), & \text{if } i = 1 \\ b_9(shiftRotation(x^{d,i})), & \text{if } i = 2, x_j^d \in [-100,100] \\ b_{10}(shiftRotation(x^{d,i})), & \text{if } i = 3 \end{cases} \end{cases}$$

$$T_2 : \begin{cases} \text{Formulation model: } F_4(x), m = 3, n = 51, K = 2, L = 49 \\ \text{ShapeFunction: } H_3(x^p) \\ \text{LandscapeFunction: } g_i(x^{d,i}) = \begin{cases} b_5(shiftRotation(x^{d,i})), & \text{if } i = 1 \\ b_7(shiftRotation(x^{d,i})), & \text{if } i = 2, x_j^d \in [-100,100] \\ b_8(shiftRotation(x^{d,i})), & \text{if } i = 3 \end{cases} \end{cases}$$

Different shift vectors and rotation matrixes are used in $T_1$ and $T_2$

5) **ETMOF5** is a 2-task optimization problem, where both task $T_1$ and task $T_2$ are 3-objective problems with an inverted linear PF and an inverted sphere PF, respectively. $T_1$ and $T_2$ are defined as follows:

$$T_1 : \begin{cases} \text{Formulation model: } F_4(x), m = 3, n = 51, K = 2, L = 49 \\ \text{ShapeFunction: } H_4(x^p) \\ \text{LandscapeFunction: } g_i(x^{d,i}) = \begin{cases} b_4(shiftRotation(x^{d,i})), & \text{if } i = 1 \\ b_6(shiftRotation(x^{d,i})), & \text{if } i = 2, x_j^d \in [-100,100] \\ b_8(shiftRotation(x^{d,i})), & \text{if } i = 3 \end{cases} \end{cases}$$

$$T_2 : \begin{cases} \text{Formulation model: } F_4(x), m = 3, n = 51, K = 2, L = 49 \\ \text{ShapeFunction: } H_6(x^p) \\ \text{LandscapeFunction: } g_i(x^{d,i}) = \begin{cases} b_1(shiftRotation(x^{d,i})), & \text{if } i = 1 \\ b_6(shiftRotation(x^{d,i})), & \text{if } i = 2, x_j^d \in [-100,100] \\ b_9(shiftRotation(x^{d,i})), & \text{if } i = 3 \end{cases} \end{cases}$$

Different shift vectors and rotation matrixes are used in $T_1$ and $T_2$

6) **ETMOF6** is a 2-task optimization problem. Task $T_1$ is a 2-objective problems with a circle PF and Task $T_2$ is a 2-objective problems with a convex PF, respectively. $T_1$ and $T_2$ are defined as follow:

$$T_1: \begin{cases} \text{Formulation model: } F_1(x), m=2, n=50, K=1, L=49 \\ \text{ShapeFunction: } H_5(x^p) \\ \text{LandscapeFunction: } g(x^d) = b_{13}(shiftRotation(x^d)), x_j^d \in [-100,100] \end{cases}$$

$$T_2: \begin{cases} \text{Formulation model: } F_4(x), m=2, n=50, K=1, L=49 \\ \text{ShapeFunction: } H_2(x^p) \\ \text{LandscapeFunction: } g_i(x^{d,i}) = \begin{cases} b_{11}(shiftRotation(x^{d,i})), & \text{if } i=1 \\ b_{12}(shiftRotation(x^{d,i})), & \text{if } i=2 \end{cases}, x_j^d \in [-100,100] \end{cases}$$

Different shift vectors and rotation matrixes are used in $T_1$ and $T_2$.

7) **ETMOF7** is a 3-task optimization problem. Task $T_1$ is a 2-objective problems with a convex PF, task $T_2$ is a 2-objective problems with a concave PF, and task $T_3$ is a 2-objective problem with a linear PF, respectively. $T_1$, $T_2$ and $T_3$ are defined as follow:

$$T_1: \begin{cases} \text{Formulation model: } F_2(x), m=2, n=50, K=1, L=49 \\ \text{ShapeFunction: } H_1(x^p) \\ \text{LandscapeFunction: } g(x^d) = L_2(rotation(x^d), y_1), x_j^d \in [-50,50] \end{cases}$$

$$T_2: \begin{cases} \text{Formulation model: } F_3(x), m=2, n=50, K=1, L=49 \\ \text{ShapeFunction: } H_2(x^p) \\ \text{LandscapeFunction: } g(x^d) = L_3(rotation(x^d), y_1), x_j^d \in [-50,50] \end{cases}$$

$$T_3: \begin{cases} \text{Formulation model: } F_2(x), m=2, n=50, K=1, L=49 \\ \text{ShapeFunction: } H_2(x^p) \\ \text{LandscapeFunction: } g(x^d) = L_8(rotation(x^d), y_1), x_j^d \in [-50,50] \end{cases}$$

Different rotation matrixes are used in $T_1$, $T_2$, and $T_3$.

8) **ETMOF8** is a 3-task optimization problem, where each task is a 3-objective problem with an inverted linear PF. $T_1$, $T_2$ and $T_3$ are defined as follows:

$$T_1: \begin{cases} \text{Formulation model: } F_7(x), m=3, n=50, K=7, L=43 \\ \text{ShapeFunction: } H_4(x^p) \\ \text{LandscapeFunction: } g_i(x^{d,i,j}) = \begin{cases} b_1(x^{d,i,j}), & \text{if } i=1 \\ b_5(x^{d,i,j}), & \text{if } i=2 \end{cases}, x_j^d \in [-10,10] \end{cases}$$

$$T_2 : \begin{cases} \text{Formulation model: } F_7(x), m=3, n=50, K=7, L=43 \\ \text{ShapeFunction: } H_4(x^p) \\ \text{LandscapeFunction: } g_i(x^{d,i,j}) = \begin{cases} b_1(x^{d,i,j}), & \text{if } i=1 \\ b_6(x^{d,i,j}), & \text{if } i=2 \end{cases}, x_j^d \in [-10,10] \end{cases}$$

$$T_3 : \begin{cases} \text{Formulation model: } F_7(x), m=3, n=50, K=7, L=43 \\ \text{ShapeFunction: } H_4(x^p) \\ \text{LandscapeFunction: } g_i(x^{d,i,j}) = \begin{cases} b_1(x^{d,i,j}), & \text{if } i=1 \\ b_8(x^{d,i,j}), & \text{if } i=2 \end{cases}, x_j^d \in [-10,10] \end{cases}$$

9) **ETMOF9** is a 2-task optimization problem. Task $T_1$ is a 5-objective problem with an inverted linear PF and Task $T_2$ is a 5-objective problems with an inverted sphere PF, respectively. $T_1$ and $T_2$ are defined as follow

$$T_1 : \begin{cases} \text{Formulation model: } F_1(x), m=5, n=25, K=4, L=21 \\ \text{ShapeFunction: } H_4(x^p) \\ \text{LandscapeFunction: } g(x^d) = b_{13}(x^d), x_j^d \in [-10,10] \end{cases}$$

$$T_2 : \begin{cases} \text{Formulation model: } F_1(x), m=5, n=53, K=4, L=49 \\ \text{ShapeFunction: } H_6(x^p) \\ \text{LandscapeFunction: } g(x^d) = b_{13}(shift(x^d)), x_j^d \in [-100,100] \end{cases}$$

10) **ETMOF10** is a 2-task optimization problem. Both task $T_1$ and Task $T_2$ are 8-objective problems with irregular concave PFs. $T_1$ and $T_2$ are defined as follow:

$$T_1 : \begin{cases} \text{Formulation model: } F_2(x), m=8, n=56, K=7, L=49 \\ \text{ShapeFunction: } H_9(x^p) \\ \text{LandscapeFunction: } g_i(x^{d,i}) = b_{14}(rotation(x^{d,i})), x_j^d \in [-20,20] \end{cases}$$

$$T_2 : \begin{cases} \text{Formulation model: } F_3(x), m=8, n=56, K=7, L=49 \\ \text{ShapeFunction: } H_9(x^p) \\ \text{LandscapeFunction: } g_i(x^{d,i}) = b_9(rotation(x^{d,i})), x_j^d \in [-10,10] \end{cases}$$

Different rotation matrixes are used in $T_1$ and $T_2$

11) **ETMOF11** is a 2-task optimization problem. Both task $T_1$ and Task $T_2$ are 10-objective problems with convex PFs. $T_1$ and $T_2$ are defined as follow:

$$T_1 : \begin{cases} \text{Formulation model: } F_2(x), m=10, n=50, K=9, L=41 \\ \text{ShapeFunction: } H_7(x^p) \\ \text{LandscapeFunction: } g_i(x^{d,i}) = L_6(x^{d,i}), x_j^d \in [-20,20] \end{cases}$$

$$T_2 : \begin{cases} \text{Formulation model: } F_3(x), m=10, n=50, K=9, L=41 \\ \text{ShapeFunction: } H_7(x^p) \\ \text{LandscapeFunction: } g_i(x^{d,i}) = L_7(x^{d,i}), x_j^d \in [-10,10] \end{cases}$$

12) **ETMOF12** is a 3-task optimization problem, where task $T_1$ is a 5-objective problem with a linear PF, Task $T_2$ is 8-objective problem with a sphere PF, and $T_3$ is a 10-objective problem with convex PF. $T_1$, $T_2$, $T_3$ and are defined as follow:

$$T_1 : \begin{cases} \text{Formulation model: } F_2(x), m=5, n=53, K=4, L=49 \\ \text{ShapeFunction: } H_3(x^p) \\ \text{LandscapeFunction: } g_i(x^{d,i}) = \begin{cases} L_4(rotation(x^{d,i}), y_1), & \text{if } i \bmod 2 = 1 \\ L_5(rotation(x^{d,i}), y_1), & \text{otherwise} \end{cases}, x_j^d \in [-10,10] \end{cases}$$

$$T_2 : \begin{cases} \text{Formulation model: } F_2(x), m=8, n=56, K=7, L=49 \\ \text{ShapeFunction: } H_5(x^p) \\ \text{LandscapeFunction: } g_i(x^{d,i}) = L_6(rotation(x^{d,i}), y_1, y_2), x_j^d \in [-10,10] \end{cases}$$

$$T_3 : \begin{cases} \text{Formulation model: } F_2(x), m=10, n=58, K=9, L=49 \\ \text{ShapeFunction: } H_7(x^p) \\ \text{LandscapeFunction: } g_i(x^{d,i}) = L_7(rotation(x^{d,i}), y_1, y_2), x_j^d \in [-10,10] \end{cases}$$

Different rotation matrixes are used in $T_1$, $T_2$, and $T_3$.

13) **ETMOF13** is a 3-task optimization problem, where task $T_1$ is a 5-objective problem with a degenerated PF, Task $T_2$ is 8-objective problem with a sphere PF, and $T_3$ is a 10-objective problem with irregular concave PF. $T_1$, $T_2$, $T_3$ and are defined as follow:

$$T_1 : \begin{cases} \text{Formulation model: } F_1(x), m=5, n=53, K=4, L=49 \\ \text{ShapeFunction: } H_8(x^p) \\ \text{LandscapeFunction: } g(x^d) = b_9(rotation(x^d)), x_j^d \in [-10,10] \end{cases}$$

$$T_2 : \begin{cases} \text{Formulation model: } F_3(x), m=8, n=56, K=7, L=49 \\ \text{ShapeFunction: } H_5(x^p) \\ \text{LandscapeFunction: } g_i(x^{d,i}) = L_2(rotation(x^{d,i}), y_1, y_2), x_j^d \in [-10,10] \end{cases}$$

$$T_3 : \begin{cases} \text{Formulation model: } F_3(x), m=10, n=58, K=9, L=49 \\ \text{ShapeFunction: } H_9(x^p) \\ \text{LandscapeFunction: } g_i(x^{d,i}) = L_3(rotation(x^{d,i}), y_1, y_2), x_j^d \in [-10,10] \end{cases}$$

Different rotation matrixes are used in $T_1$, $T_2$, and $T_3$

14) **ETMOF14** is a 3-task optimization problem, where task $T_1$ is a 5-objective problem with a disconnected PF, Task $T_2$ is 8-objective problem with a disconnected PF, and $T_3$ is a 10-objective problem with a disconnected PF. $T_1$, $T_2$, $T_3$ and are defined as follow:

$$T_1 : \begin{cases} \text{Formulation model: } F_5(x), m=5, n=53, K=4, L=49 \\ \text{ShapeFunction: } H_{10}(x^p) \\ \text{LandscapeFunction: } g_i(x^{d,i}) = \begin{cases} b_1(rotation(x^{d,i})), & \text{if } i \bmod 2 = 1 \\ b_5(rotation(x^{d,i})), & \text{otherwise} \end{cases}, x_j^d \in [-1,1] \end{cases}$$

$$T_2 : \begin{cases} \text{Formulation model: } F_5(x), m=8, n=56, K=7, L=49 \\ \text{ShapeFunction: } H_{10}(x^p) \\ \text{LandscapeFunction: } g_i(x^{d,i}) = \begin{cases} b_{14}(rotation(x^{d,i})), & \text{if } i \bmod 2 = 1 \\ b_9(rotation(x^{d,i})), & \text{otherwise} \end{cases}, x_j^d \in [-1,1] \end{cases}$$

$$T_3 : \begin{cases} \text{Formulation model: } F_4(x), m=10, n=58, K=9, L=49 \\ \text{ShapeFunction: } H_{10}(x^p) \\ \text{LandscapeFunction: } g_i(x^{d,i}) = \begin{cases} b_1(rotation(x^{d,i})), & \text{if } i \bmod 2 = 1 \\ b_{14}(rotation(x^{d,i})), & \text{otherwise} \end{cases}, x_j^d \in [-1,1] \end{cases}$$

Different rotation matrixes are used in $T_1$, $T_2$, and $T_3$

15) **ETMOF15** is a 2-task optimization problem, where each task is a 10-objective problem with a sphere PF. $T_1$ and $T_2$ are defined as follows:

$$T_1 : \begin{cases} \text{Formulation model: } F_6(x), m=10, n=99, K=28, L=71 \\ \text{ShapeFunction: } H_5(x^p) \\ \text{LandscapeFunction: } g_i(x^{d,i,j}) = b_1(x^{d,i,j}), x_j^d \in [-10,10] \end{cases}$$

$$T_2 : \begin{cases} \text{Formulation model: } F_7(x), m=10, n=99, K=28, L=71 \\ \text{ShapeFunction: } H_5(x^p) \\ \text{LandscapeFunction: } g_i(x^{d,i,j}) = b_{14}(x^{d,i,j}), x_j^d \in [-10,10] \end{cases}$$

16) **ETMOF16** is a 2-task optimization problem, where each task is a 5-objective problem with a linear PF. $T_1$ and $T_2$ are defined as follows:

$$T_1 : \begin{cases} \text{Formulation model: } F_6(x), m=5, n=80, K=13, L=67 \\ \text{ShapeFunction: } H_3(x^p) \\ \text{LandscapeFunction: } g_i(x^{d,i,j}) = \begin{cases} b_5(x^{d,i,j}), & \text{if } i=1 \\ b_8(x^{d,i,j}), & \text{if } i=2 \end{cases}, x_j^d \in [-10,10] \end{cases}$$

$$T_2 : \begin{cases} \text{Formulation model: } F_7(x), m=5, n=80, K=13, L=67 \\ \text{ShapeFunction: } H_3(x^p) \\ \text{LandscapeFunction: } g_i(x^{d,i,j}) = \begin{cases} b_5(x^{d,i,j}), & \text{if } i=1 \\ b_9(x^{d,i,j}), & \text{if } i=2 \end{cases}, x_j^d \in [-10,10] \end{cases}$$

17) **ETMOF17** is a 2-task optimization problem, where task $T_1$ is a 3-objective large-scale problem with a linear PF, and task $T_2$ is a 3-objective large-scale problem with an inverted linear PF. $T_1$ and $T_2$ are defined as follows:

$$T_1 : \begin{cases} \text{Formulation model: } F_6(x), m=3, n=256, K=11, L=245 \\ \text{ShapeFunction: } H_3(x^p) \\ \text{LandscapeFunction: } g(x^{d,i,j}) = b_1(Lg_1(x^{d,i,j}, y_1)), x_j^d \in [-10,10] \end{cases}$$

$$T_2 : \begin{cases} \text{Formulation model: } F_6(x), m=3, n=256, K=11, L=245 \\ \text{ShapeFunction: } H_4(x^p) \\ \text{LandscapeFunction: } g(x^{d,i,j}) = b_1(Lg_2(x^{d,i,j}, y_1)), x_j^d \in [-10,10] \end{cases}$$

where $Lg_1(x^{d,i,j}, y_1)$ and $Lg_2(x^{d,i,j}, y_1)$ represent the linear linkage operator and nonlinear linkage operator, respectively, which are the same as proposed in [13].

18) **ETMOF18** is a 2-task optimization problem, where task $T_1$ is a 2-objective large-scale problem with a circle PF, and task $T_2$ is a 2-objective large-scale problem with an convex PF. $T_1$ and $T_2$ are defined as follows:

$$T_1 : \begin{cases} \text{Formulation model: } F_7(x), m=2, n=512, K=6, L=506 \\ \text{ShapeFunction: } H_5(x^p) \\ \text{LandscapeFunction: } g(x^{d,i,j}) = b_1(Lg_1(x^{d,i,j}, y_1)), x_j^d \in [-10,10] \end{cases}$$

$$T_2 : \begin{cases} \text{Formulation model: } F_7(x), m=2, n=512, K=6, L=506 \\ \text{ShapeFunction: } H_7(x^p) \\ \text{LandscapeFunction: } g(x^{d,i,j}) = b_{14}(Lg_1(x^{d,i,j}, y_1)), x_j^d \in [-10,10] \end{cases}$$

19) **ETMOF19** is a 2-task optimization problem, where both task $T_1$ and task $T_2$ are 2-objective large-scale problems with circle PFs. $T_1$ and $T_2$ are defined as follows:

$$T_1 : \begin{cases} \text{Formulation model: } F_6(x), m=2, n=1024, K=6, L=1018 \\ \text{ShapeFunction: } H_3(x^p) \\ \text{LandscapeFunction: } g_i(x^{d,i}) = \begin{cases} b_1(Lg_1(x^{d,i,j})), & \text{if } i=1 \\ b_5(Lg_1(x^{d,i,j})), & \text{if } i=2 \end{cases}, x_j^d \in [-10,10] \end{cases}$$

$$T_2 : \begin{cases} \text{Formulation model: } F_6(x), m=2, n=1024, K=6, L=1018 \\ \text{ShapeFunction: } H_3(x^p) \\ \text{LandscapeFunction: } g_i(x^{d,i}) = \begin{cases} b_1(Lg_1(x^{d,i,j})), & \text{if } i=1 \\ b_9(Lg_1(x^{d,i,j})), & \text{if } i=2 \end{cases}, x_j^d \in [-10,10] \end{cases}$$

20) **ETMOF20** is a 3-task optimization problem, where task $T_1$, task $T_2$, and task $T_3$ are 2-objective large-scale problems with convex, linear, and circle PFs, respectively. $T_1$, $T_2$ and $T_3$ are defined as follows:

$$T_1 : \begin{cases} \text{Formulation model: } F_7(x), m=2, n=256, K=6, L=250 \\ \text{ShapeFunction: } H_2(x^p) \\ \text{LandscapeFunction: } g_i(x^{d,i}) = b_1(Lg_2(x^{d,i,j})), x_j^d \in [-10,10] \end{cases}$$

$$T_2 : \begin{cases} \text{Formulation model: } F_7(x), m=2, n=512, K=6, L=506 \\ \text{ShapeFunction: } H_3(x^p) \\ \text{LandscapeFunction: } g_i(x^{d,i}) = b_{14}(Lg_2(x^{d,i,j})), x_j^d \in [-10,10] \end{cases}$$

$$T_3: \begin{cases} \text{Formulation model: } F_7(x), m = 2, n = 1024, K = 6, L = 1018 \\ \text{ShapeFunction: } H_5(x^p) \\ \text{LandscapeFunction: } g_i(x^{d,i}) = \begin{cases} b_1(Lg_2(x^{d,i,j})), & \text{if } i = 1 \\ b_{14}(Lg_2(x^{d,i,j})), & \text{if } i = 2 \end{cases}, x_j^d \in [-10, 10] \end{cases}$$

21) **ETMOF21** is a 3-task optimization problem, where task $T_1$, task $T_2$, and task $T_3$ are 3-objective large-scale problems with sphere PFs. $T_1$, $T_2$ and $T_3$ are defined as follows:

$$T_1: \begin{cases} \text{Formulation model: } F_6(x), m = 3, n = 512, K = 11, L = 501 \\ \text{ShapeFunction: } H_5(x^p) \\ \text{LandscapeFunction: } g_i(x^{d,i,j}) = \begin{cases} b_1(Lg_2(x^{d,i,j})), & \text{if } i = 1 \\ b_5(Lg_2(x^{d,i,j})), & \text{if } i = 2, x_j^d \in [-10, 10] \\ b_6(Lg_2(x^{d,i,j})), & \text{if } i = 3 \end{cases} \end{cases}$$

$$T_2: \begin{cases} \text{Formulation model: } F_6(x), m = 3, n = 512, K = 11, L = 501 \\ \text{ShapeFunction: } H_5(x^p) \\ \text{LandscapeFunction: } g_i(x^{d,i,j}) = \begin{cases} b_{14}(Lg_2(x^{d,i,j})), & \text{if } i = 1 \\ b_5(Lg_2(x^{d,i,j})), & \text{if } i = 2, x_j^d \in [-10, 10] \\ b_8(Lg_2(x^{d,i,j})), & \text{if } i = 3 \end{cases} \end{cases}$$

$$T_3: \begin{cases} \text{Formulation model: } F_6(x), m = 3, n = 512, K = 11, L = 501 \\ \text{ShapeFunction: } H_5(x^p) \\ \text{LandscapeFunction: } g_i(x^{d,i,j}) = \begin{cases} b_1(Lg_2(x^{d,i,j})), & \text{if } i = 1 \\ b_9(Lg_2(x^{d,i,j})), & \text{if } i = 2, x_j^d \in [-10, 10] \\ b_6(Lg_2(x^{d,i,j})), & \text{if } i = 3 \end{cases} \end{cases}$$

22) **ETMOF22** is a 3-task optimization problem, where task $T_1$, task $T_2$, and task $T_3$ are 3-objective large-scale problems with inverted linear PFs. $T_1$, $T_2$ and $T_3$ are defined as follows:

$$T_1: \begin{cases} \text{Formulation model: } F_6(x), m = 3, n = 256, K = 11, L = 245 \\ \text{ShapeFunction: } H_4(x^p) \\ \text{LandscapeFunction: } g_i(x^{d,i,j}) = \begin{cases} b_5(Lg_1(x^{d,i,j})), & \text{if } i = 1 \\ b_6(Lg_1(x^{d,i,j})), & \text{if } i = 2, x_j^d \in [-10, 10] \\ b_9(Lg_1(x^{d,i,j})), & \text{if } i = 3 \end{cases} \end{cases}$$

$$T_2: \begin{cases} \text{Formulation model: } F_7(x), m = 3, n = 512, K = 11, L = 501 \\ \text{ShapeFunction: } H_4(x^p) \\ \text{LandscapeFunction: } g_i(x^{d,i,j}) = \begin{cases} b_5(Lg_2(x^{d,i,j})), & \text{if } i = 1 \\ b_6(Lg_2(x^{d,i,j})), & \text{if } i = 2, x_j^d \in [-10, 10] \\ b_9(Lg_2(x^{d,i,j})), & \text{if } i = 3 \end{cases} \end{cases}$$

$$T_3 : \begin{cases} \text{Formulation model: } F_6(x), m=3, n=1024, K=11, L=1013 \\ \text{ShapeFunction: } H_4(x^p) \\ \text{LandscapeFunction: } g_i(x^{d,i,j}) = \begin{cases} b_5(Lg_2(x^{d,i,j})), & \text{if } i=1 \\ b_6(Lg_2(x^{d,i,j})), & \text{if } i=2, x_j^d \in [-10,10] \\ b_9(Lg_2(x^{d,i,j})), & \text{if } i=3 \end{cases} \end{cases}$$

23) **ETMOF23** is a 2-task optimization problem, where both task $T_1$ and task $T_2$ are 2-objective large-scale problems with linear PFs. $T_1$ and $T_2$ are defined as follows:

$$T_1 : \begin{cases} \text{Formulation model: } F_6(x), m=2, n=2048, K=6, L=2042 \\ \text{ShapeFunction: } H_3(x^p) \\ \text{LandscapeFunction: } g_i(x^{d,i,j}) = b_1(Lg_1(x^{d,i,j})), x_j^d \in [-10,10] \end{cases}$$

$$T_2 : \begin{cases} \text{Formulation model: } F_6(x), m=2, n=4096, K=6, L=4090 \\ \text{ShapeFunction: } H_3(x^p) \\ \text{LandscapeFunction: } g_i(x^{d,i,j}) = \begin{cases} b_1(Lg_1(x^{d,i,j})), & \text{if } i=1 \\ b_5(Lg_1(x^{d,i,j})), & \text{if } i=2 \end{cases}, x_j^d \in [-10,10] \end{cases}$$

24) **ETMOF24** is a 2-task optimization problem, where both task $T_1$ and task $T_2$ are 2-objective large-scale problems with circle PFs. $T_1$ and $T_2$ are defined as follows:

$$T_1 : \begin{cases} \text{Formulation model: } F_6(x), m=2, n=5000, K=6 \\ \text{ShapeFunction: } H_3(x^p) \\ \text{LandscapeFunction: } g_i(x^{d,i,j}) = \begin{cases} b_1(Lg_1(x^{d,i,j})), & \text{if } i=1 \\ b_9(Lg_1(x^{d,i,j})), & \text{if } i=2 \end{cases}, x_j^d \in [-10,10] \end{cases}$$

$$T_2 : \begin{cases} \text{Formulation model: } F_6(x), m=2, n=10000, K=6 \\ \text{ShapeFunction: } H_3(x^p) \\ \text{LandscapeFunction: } g_i(x^{d,i,j}) = \begin{cases} b_5(Lg_1(x^{d,i,j})), & \text{if } i=1 \\ b_9(Lg_1(x^{d,i,j})), & \text{if } i=2 \end{cases}, x_j^d \in [-10,10] \end{cases}$$

25) **ETMOF25** is a 5-task optimization problem, where task $T_1$ to $T_5$ are defined as follows:

$$T_{k=1 \text{ to } 5} : \begin{cases} \text{Formulation model: } F_3(x), m=2, n=50, K=1, L=49 \\ \text{ShapeFunction: } H_1(x^p) \\ \text{LandscapeFunction: } g_i(x^{d,i}) = \begin{cases} L_1(rotation(x^{d,i}), y_1), x_j^d \in [-10,10] & \text{if } k=1 \\ L_2(rotation(x^{d,i}), y_1), x_j^d \in [-5,5] & \text{if } k=2 \\ L_3(rotation(x^{d,i}), y_1), x_j^d \in [-1,1] & \text{if } k=3 \\ L_5(rotation(x^{d,i}), y_1), x_j^d \in [-5,5] & \text{if } k=4 \\ L_7(rotation(x^{d,i}), y_1), x_j^d \in [-10,10] & \text{if } k=5 \end{cases} \end{cases}$$

Different rotation matrixes are used in $T_1$ to $T_5$.

26) **ETMOF26** is a 10-task optimization problem, where task $T_1$ to $T_{10}$ are defined as follows:

$$T_{k=1 \text{ to } 10}: \begin{cases} \text{Formulation model:} \begin{cases} F_2(x), & \text{if } k \bmod 2 = 0 \\ F_3(x), & \text{if } k \bmod 2 = 1 \end{cases}, m=2, n=50, K=1, L=49 \\ \text{ShapeFunction:} \begin{cases} H_1(x^p), & \text{if } k \bmod 3 = 0 \\ H_2(x^p), & \text{if } k \bmod 3 = 1 \\ H_3(x^p), & \text{if } k \bmod 3 = 2 \end{cases} \\ \text{LandscapeFunction: } g_i(x^{d,i}) = \begin{cases} L_1(rotation(x^{d,i}), y_1), x_j^d \in [-10,10], & \text{if } k \bmod 5 = 0 \\ L_2(rotation(x^{d,i}), y_1), x_j^d \in [-5,5], & \text{if } k \bmod 5 = 1 \\ L_3(rotation(x^{d,i}), y_1), x_j^d \in [-1,1], & \text{if } k \bmod 5 = 2 \\ L_4(rotation(x^{d,i}), y_1), x_j^d \in [-5,5], & \text{if } k \bmod 5 = 3 \\ L_7(rotation(x^{d,i}), y_1), x_j^d \in [-10,10], & \text{if } k \bmod 5 = 4 \end{cases} \end{cases}$$

Different rotation matrixes are used in $T_1$ to $T_{10}$.

27) **ETMOF27** is a 10-task optimization problem, where task $T_1$ to $T_{10}$ are defined as follows:

$$T_{k=1 \text{ to } 10}: \begin{cases} \text{Formulation model:} \begin{cases} F_6(x), & \text{if } k/5 \leq 1 \\ F_7(x), & \text{otherwisw} \end{cases}, m=3, n=50, K=7, L=43 \\ \text{ShapeFunction:} H_3(x^p) \\ \text{LandscapeFunction: } g_i(x^{d,i,j}) = \begin{cases} b_5(z), x_j^d \in [-10,10], & \text{if } k \bmod 5 = 0 \\ b_6(z), x_j^d \in [-10,10], & \text{if } k \bmod 5 = 1 \\ b_7(z), x_j^d \in [-10,10], & \text{if } k \bmod 5 = 2 \\ b_8(z), x_j^d \in [-10,10], & \text{if } k \bmod 5 = 3 \\ b_9(z), x_j^d \in [-10,10], & \text{if } k \bmod 5 = 4 \end{cases}, z = \begin{cases} Lg_1(x^{d,i,j}), & \text{if } k/5 \leq 1 \\ Lg_2(x^{d,i,j}), & \text{otherwisw} \end{cases} \end{cases}$$

Different rotation matrixes are used in $T_1$ to $T_{10}$.

28) **ETMOF28** is a 20-task optimization problem, where task $T_1$ to $T_{20}$ are defined as follows:

$$T_{k=1 \text{ to } 20}: \begin{cases} \text{Formulation model:} \begin{cases} F_2(x), & \text{if } k \bmod 2 = 0 \\ F_3(x), & \text{if } k \bmod 2 = 1 \end{cases}, m=3, n=51, K=2, L=49 \\ \text{ShapeFunction:} \begin{cases} H_3(x^p), & \text{if } k \bmod 3 = 0 \\ H_5(x^p), & \text{if } k \bmod 3 = 1 \\ H_7(x^p), & \text{if } k \bmod 3 = 2 \end{cases} \\ \text{LandscapeFunction: } g_i(x^{d,i}) = \begin{cases} b_1(rotation(x^{d,i})), x_j^d \in [-50,50], & \text{if } k \bmod 5 = 0 \\ b_5(rotation(x^{d,i})), x_j^d \in [-10,10], & \text{if } k \bmod 5 = 1 \\ b_6(rotation(x^{d,i})), x_j^d \in [-20,20], & \text{if } k \bmod 5 = 2 \\ b_8(rotation(x^{d,i})), x_j^d \in [-30,30], & \text{if } k \bmod 5 = 3 \\ b_9(rotation(x^{d,i})), x_j^d \in [-40,40], & \text{if } k \bmod 5 = 4 \end{cases} \end{cases}$$

Different rotation matrixes are used in $T_1$ to $T_{20}$.

29) **ETMOF29** is a 30-task optimization problem, where task $T_1$ to $T_{30}$ are defined as follows:

$$T_{k=1 \text{ to } 30} : \begin{cases} \text{Formulation model:} F_1(x), m = 3, n = 51, K = 2, L = 49 \\ \text{ShapeFunction:} \begin{cases} H_4(x^p), & \text{if } k \bmod 2 = 0 \\ H_6(x^p), & \text{if } k \bmod 2 = 1 \end{cases} \\ \text{LandscapeFunction: } g_i(x) = \begin{cases} b_1(shiftRotation(x)), x_j^d \in [-50, 50], & \text{if } k \bmod 3 = 0 \\ b_9(shiftRotation(x)), x_j^d \in [-100, 100], & \text{if } k \bmod 3 = 1 \\ b_6(shiftRotation(x)), x_j^d \in [-0.5, 0.5], & \text{if } k \bmod 3 = 2 \end{cases} \end{cases}$$

Different shift vectors and rotation matrixes are used in $T_1$ to $T_{30}$.

30) **ETMOF30** is a 40-task optimization problem, where task $T_1$ to $T_{40}$ are defined as follows:

$$T_{k=1 \text{ to } 40} : \begin{cases} \text{Formulation model:} F_1(x), m = 2, n = 50, K = 1, L = 49 \\ \text{ShapeFunction:} H_1(x^p) \\ \text{LandscapeFunction: } g(x^d) = \begin{cases} b_5(shiftRotation(x^d)), x_j^d \in [-50, 50], & \text{if } k \bmod 5 = 0 \\ b_6(shiftRotation(x^d)), x_j^d \in [-50, 50], & \text{if } k \bmod 5 = 1 \\ b_9(shiftRotation(x^d)), x_j^d \in [-50, 50], & \text{if } k \bmod 5 = 2 \\ b_7(shiftRotation(x^d)), x_j^d \in [-100, 100], & \text{if } k \bmod 5 = 3 \\ b_8(shiftRotation(x^d)), x_j^d \in [-0.5, 0.5], & \text{if } k \bmod 5 = 4 \end{cases} \end{cases}$$

Different shift vectors and rotation matrixes are used in $T_1$ to $T_{40}$.

31) **ETMOF31** is a 50-task optimization problem, where task $T_1$ to $T_{50}$ are defined as follows:

$$T_{k=1 \text{ to } 50} : \begin{cases} \text{Formulation model:} F_3(x), m = 2, n = 50, K = 1, L = 49 \\ \text{ShapeFunction:} H_2(x^p) \\ \text{LandscapeFunction: } g_i(x^{d,i}) = \begin{cases} L_1(rotation(x^{d,i}), y_1), x_j^d \in [-60, 60], & \text{if } k \bmod 6 = 0 \\ L_2(rotation(x^{d,i}), y_1), x_j^d \in [-50, 50], & \text{if } k \bmod 6 = 1 \\ L_3(rotation(x^{d,i}), y_1), x_j^d \in [-40, 40], & \text{if } k \bmod 6 = 2 \\ L_4(rotation(x^{d,i}), y_1), x_j^d \in [-30, 30], & \text{if } k \bmod 6 = 3 \\ L_5(rotation(x^{d,i}), y_1), x_j^d \in [-20, 20], & \text{if } k \bmod 6 = 4 \\ L_7(rotation(x^{d,i}), y_1), x_j^d \in [-10, 10], & \text{if } k \bmod 6 = 5 \end{cases} \end{cases}$$

Different rotation matrixes are used in $T_1$ to $T_{50}$.

32) **ETMOF32** is a 28-task optimization problem, where task $T_1$ to $T_{28}$ are defined as follows:

$$T_{k=1 \text{ to } 28} : \begin{cases} \text{Formulation model:} F_6(x), m = 3, n = 80, K = 7, L = 73 \\ \text{ShapeFunction:} H_5(x^p) \\ \text{LandscapeFunction: } g_i(x^{d,i,j}) = \begin{cases} b_k(Lg_1(x^{d,i,j})), x_j^d \in [-10, 10], & \text{if } k \leq 10 \\ b_{k-14}(Lg_2(x^{d,i,j})), x_j^d \in [-10, 10], & \text{otherwise} \end{cases} \end{cases}$$

33) **ETMOF33** is a 2-task optimization problem, where task $T_1$ and task $T_2$ are both dynamic 2-objective optimization problems. Here, $T_1$ and $T_2$ are the same with dMOP2[14] and ZJZ [15] with large-scale of variables, respectively, which are defined as follows:

$$T_1: \begin{cases} \min \begin{cases} f_1(\mathbf{x}) = x^p \\ f_2(\mathbf{x}) = (1+g(\mathbf{x}^d))\left(1-\left(\dfrac{x^p}{g(\mathbf{x}^d)}\right)^{H(t)}\right) \end{cases}, g(\mathbf{x}^d) = b_1(\mathbf{x}^d - G(t)), m=2, n=256, K=1 \\ H(t) = 0.75\sin(0.5\pi t)+1.25, G(t) = \|\sin(0.5\pi t)\|, t = \dfrac{1}{n_t}\left\lfloor \dfrac{\tau}{\tau_t} \right\rfloor, n_t=10, \tau_t=20, x_i^d \in [-1,1] \end{cases}$$

$$T_2: \begin{cases} \min \begin{cases} f_1(\mathbf{x}) = x^p \\ f_2(\mathbf{x}) = (1+g(\mathbf{x}^d))\left(1-\left(\dfrac{x^p}{g(\mathbf{x}^d)}\right)^{H(t)}\right) \end{cases}, g(\mathbf{x}^d) = b_1\left(x_i^d - G(t) - (x^p)^{H(t)}\right), m=2, n=256, K=1 \\ G(t) = \sin(0.5\pi t), H(t) = 1.5+G(t), t = \dfrac{1}{n_t}\left\lfloor \dfrac{\tau}{\tau_t} \right\rfloor, n_t=10, \tau_t=20, x_i^d \in [-1,2] \end{cases}$$

where $\tau$ indicates the generation counter, $\tau_t$ indicates the frequency of environment change in each run, $n_t$ used to control the severity of the environment change, $t$ is the time instant used to dynamically evaluate the functions.

34) **ETMOF34** is a 2-task optimization problem, where task $T_1$ and task $T_2$ are both dynamic 2-objective optimization problem. Here, $T_1$ is the same with DF2[16] and $T_2$ is a modified DF2, which are defined as follows:

$$T_1: \begin{cases} \min \begin{cases} f_1(\mathbf{x}) = x_r \\ f_2(\mathbf{x}) = (1+g(\mathbf{x}^d))\left(1-\sqrt{f_1/g}\right) \end{cases}, g(\mathbf{x}^d) = b_1(\mathbf{x}^d - G(t)), m=2, n=50, K=1, L=49 \\ G(t) = \|\sin(0.5\pi t)\|, r = 1+\lfloor (n-1)G(t) \rfloor, t = \dfrac{1}{n_t}\left\lfloor \dfrac{\tau}{\tau_t} \right\rfloor, n_t=10, \tau_t=20, x_i^d \in [0,1] \end{cases}$$

$$T_2: \begin{cases} \min \begin{cases} f_1(\mathbf{x}) = x_r \\ f_2(\mathbf{x}) = (1+g(\mathbf{x}^d))\left(1-\sqrt{f_1/g}\right) \end{cases}, g(\mathbf{x}^d) = b_9(\mathbf{x}^d - G(t)), m=2, n=50, K=1, L=49 \\ G(t) = \|\sin(0.5\pi t)\|, r = 1+\lfloor (n-1)G(t) \rfloor, t = \dfrac{1}{n_t}\left\lfloor \dfrac{\tau}{\tau_t} \right\rfloor, n_t=10, \tau_t=20, x_i^d \in [0,1] \end{cases}$$

35) **ETMOF35** is a 2-task optimization problem, where task $T_1$ and task $T_2$ are both dynamic 2-objective optimization problem. Here, $T_1$ and $T_2$ are the same with DF5 and DF6 [16] with large-scale of variables, respectively, which are defined as follows:

$$T_1: \begin{cases} \min \begin{cases} f_1(\mathbf{x}) = g(\mathbf{x}^d)\left(x^p + 0.02\sin(w_t\pi x^p)\right) \\ f_2(\mathbf{x}) = g(\mathbf{x}^d)\left(1 - x^p + 0.02\sin(w_t\pi x^p)\right) \end{cases}, g(\mathbf{x}^d) = 1+\sum_{i=1}^{|\mathbf{x}^d|}\left((x_i^d) - G(t)\right)^2 \\ G(t) = \sin(0.5\pi t), w_t = \lfloor 10G(t) \rfloor, m=2, n=512, K=1 \\ t = \dfrac{1}{n_t}\left\lfloor \dfrac{\tau}{\tau_t} \right\rfloor, n_t=10, \tau_t=20, x_i^d \in [-1,1] \end{cases}$$

$$T_2 : \begin{cases} \min \begin{cases} f_1(\mathbf{x}) = g(\mathbf{x}^d)\left[(x^p + 0.1\sin(3\pi x^p))\right]^{\alpha_t} \\ f_2(\mathbf{x}) = g(\mathbf{x}^d)\left[(1 - x^p + 0.1\sin(3\pi x^p))\right]^{\alpha_t} \end{cases}, g(\mathbf{x}^d) = 1 + \sum_{i=1}^{|\mathbf{x}^d|}\left(\|G(t)\|(y_i)^2 - 10\cos(2\pi y_i) + 10\right) \\ y_i = x_d^i - G(t), G(t) = \sin(0.5\pi t), \alpha_t = 0.2 + 0.28\|G(t)\| \\ m = 2, n = 512, K = 1, t = \dfrac{1}{n_t}\left\lfloor \dfrac{\tau}{\tau_t} \right\rfloor, n_t = 10, \tau_t = 20, x_i^d \in [-1,1] \end{cases}$$

36) **ETMOF36** is a 2-task optimization problem, where task $T_1$ and task $T_2$ are both dynamic 2-objective optimization problem. Here, $T_1$ and $T_2$ are the same with DF8 and DF6 [16] with large-scale of variables, respectively, which are defined as follows:

$$T_1 : \begin{cases} \min \begin{cases} f_1(\mathbf{x}) = g(\mathbf{x}^d)(x^p + 0.1\sin(3\pi x^p)) \\ f_2(\mathbf{x}) = g(\mathbf{x}^d)\left[1 - x^p + 0.1\sin(3\pi x^p)\right]^{\alpha_t} \end{cases}, g(\mathbf{x}^d) = 1 + \sum_{i=1}^{|\mathbf{x}^d|}\left(x_i^d - \dfrac{G(t)\sin(4\pi(x^p)^{\beta_t})}{1 + \|G(t)\|}\right)^2 \\ \alpha_t = 2.25 + 2\cos(2\pi t), \beta_t = 100G^2(t), G(t) = \sin(0.5\pi t), \\ m = 2, n = 5000, K = 1, t = \dfrac{1}{n_t}\left\lfloor \dfrac{\tau}{\tau_t} \right\rfloor, n_t = 10, \tau_t = 20, x_i^d \in [-1,1] \end{cases}$$

$$T_2 : \begin{cases} \min \begin{cases} f_1(\mathbf{x}) = g(\mathbf{x}^d)\left[x^p + 0.1\sin(3\pi x^p)\right]^{\alpha_t} \\ f_2(\mathbf{x}) = g(\mathbf{x}^d)\left[1 - x^p + 0.1\sin(3\pi x^p)\right]^{\alpha_t} \end{cases}, g(\mathbf{x}^d) = 1 + \sum_{i=1}^{|\mathbf{x}^d|}\left(\|G(t)\|(y_i)^2 - 10\cos(2\pi y_i) + 10\right) \\ y_i = x_d^i - G(t), G(t) = \sin(0.5\pi t), \alpha_t = 0.2 + 0.28\|G(t)\| \\ m = 2, n = 10000, K = 1, t = \dfrac{1}{n_t}\left\lfloor \dfrac{\tau}{\tau_t} \right\rfloor, n_t = 10, \tau_t = 20, x_i^d \in [-1,1] \end{cases}$$

37) **ETMOF37** is a 2-task optimization problem, where task $T_1$ and task $T_2$ are both dynamic 3-objective optimization problem. Here, $T_1$ and $T_2$ are the same with DF10 and DF11 [16], which are defined as follows:

$$T_1 : \begin{cases} \min \begin{cases} f_1(\mathbf{x}) = (1 + g(\mathbf{x}^d))\left[\sin(0.5\pi x_1^p)\right]^{H(t)} \\ f_2(\mathbf{x}) = (1 + g(\mathbf{x}^d))\left[\sin(0.5\pi x_2^p)\cos(0.5\pi x_1^p)\right]^{H(t)} \\ f_3(\mathbf{x}) = (1 + g(\mathbf{x}^d))\left[\cos(0.5\pi x_2^p)\cos(0.5\pi x_1^p)\right]^{H(t)} \end{cases}, g(\mathbf{x}^d) = b_1\left(\mathbf{x}^d - \dfrac{\sin(2\pi(x_1^p + x_2^p))}{1 + \|G(t)\|}\right) \\ H(t) = 2.25 + 2con(0.5\pi t), G(t) = \sin(0.5\pi t) \\ m = 3, n = 50, K = 2, t = \dfrac{1}{n_t}\left\lfloor \dfrac{\tau}{\tau_t} \right\rfloor, n_t = 10, \tau_t = 20, x_i^d \in [-1,1] \end{cases}$$

$$T_2 : \begin{cases} \min \begin{cases} f_1(\mathbf{x}) = (1 + g(\mathbf{x}^d))\sin(z_1) \\ f_2(\mathbf{x}) = (1 + g(\mathbf{x}^d))\sin(z_2)\cos(z_1) \\ f_3(\mathbf{x}) = (1 + g(\mathbf{x}^d))\cos(z_2)\cos(z_1) \end{cases}, g(\mathbf{x}^d) = G(t) + b_1(\mathbf{x}^d - 0.5G(t)x_1^p) \\ z_{i=1:2} = \dfrac{\pi}{6}G(t) + \left(\dfrac{\pi}{2} - \dfrac{\pi}{3}G(t)\right)x_i^p, G(t) = \|\sin(0.5\pi t)\| \\ m = 3, n = 50, K = 2, t = \dfrac{1}{n_t}\left\lfloor \dfrac{\tau}{\tau_t} \right\rfloor, n_t = 10, \tau_t = 20, x_i^d \in [0,1] \end{cases}$$

38) **ETMOF38** is a 2-task optimization problem, where task $T_1$ and task $T_2$ are both dynamic 3-objective optimization problem. Here, $T_1$ and $T_2$ are the same with DF12 and DF11 [16], which are defined as follows:

$$T_1 : \begin{cases} \min \begin{cases} f_1(\mathbf{x}) = (1+g(\mathbf{x}^d))\cos(0.5\pi x_1^d)\cos(0.5\pi x_2^d) \\ f_2(\mathbf{x}) = (1+g(\mathbf{x}^d))\cos(0.5\pi x_1^d)\sin(0.5\pi x_2^d) \\ f_3(\mathbf{x}) = (1+g(\mathbf{x}^d))\sin(0.5\pi x_1^d) \end{cases} \\ g(\mathbf{x}^d) = \sum_{i=1}^{|\mathbf{x}^d|}\left(x_i^d - \sin(tx_1^p)\right)^2 + \left\|\prod_{j=1}^{2}\sin\left(\lfloor k_t(2x_j^p-1)\rfloor \pi/2\right)\right\|, k_t=10\sin(\pi t) \\ m=3, n=50, K=2, t=\dfrac{1}{n_t}\left\lfloor\dfrac{\tau}{\tau_t}\right\rfloor, n_t=10, \tau_t=20, x_i^d \in [-1,1] \end{cases}$$

$$T_2 : \begin{cases} \min \begin{cases} f_1(\mathbf{x}) = (1+g(\mathbf{x}^d))\sin(z_1) \\ f_2(\mathbf{x}) = (1+g(\mathbf{x}^d))\sin(z_2)\cos(z_1), \ g(\mathbf{x}^d) = G(t) + b_1(\mathbf{x}^d - 0.5G(t)x_1^p) \\ f_3(\mathbf{x}) = (1+g(\mathbf{x}^d))\cos(z_2)\cos(z_1) \end{cases} \\ z_{i=1:2} = \dfrac{\pi}{6}G(t) + (\dfrac{\pi}{2} - \dfrac{\pi}{3}G(t))x_i^p, G(t) = \|\sin(0.5\pi t)\| \\ m=3, n=50, K=2, t=\dfrac{1}{n_t}\left\lfloor\dfrac{\tau}{\tau_t}\right\rfloor, n_t=10, \tau_t=20, x_i^d \in [0,1] \end{cases}$$

39) **ETMOF39** is a 3-task optimization problem, where task $T_1$ to task $T_3$ are dynamic 2-objective optimization problem. $T_1$, $T_2$, and $T_3$ are defined as follows

$$T_1 : \begin{cases} \min \begin{cases} f_1(\mathbf{x}) = (1+g(\mathbf{x}^d))(x^p + 0.02\sin(w_t\pi x^p)) \\ f_2(\mathbf{x}) = (1+g(\mathbf{x}^d))(1 - x^p + 0.02\sin(w_t\pi x^p)) \end{cases}, g(\mathbf{x}^d) = b_1(\mathbf{x}^d - G(t)) \\ G(t) = \sin(0.5\pi t), w_t = \lfloor 10G(t)\rfloor, m=2, n=50, K=1 \\ t = \dfrac{1}{n_t}\left\lfloor\dfrac{\tau}{\tau_t}\right\rfloor, n_t=10, \tau_t=20, x_i^d \in [-1,1] \end{cases}$$

$$T_2 : \begin{cases} \min \begin{cases} f_1(\mathbf{x}) = (1+g(\mathbf{x}^d))(x^p + 0.02\sin(w_t\pi x^p)) \\ f_2(\mathbf{x}) = (1+g(\mathbf{x}^d))(1 - x^p + 0.02\sin(w_t\pi x^p)) \end{cases}, g(\mathbf{x}^d) = b_5(\mathbf{x}^d - G(t)) \\ G(t) = \sin(0.5\pi t), w_t = \lfloor 10G(t)\rfloor, m=2, n=50, K=1 \\ t = \dfrac{1}{n_t}\left\lfloor\dfrac{\tau}{\tau_t}\right\rfloor, n_t=10, \tau_t=20, x_i^d \in [-1,1] \end{cases}$$

$$T_3 : \begin{cases} \min \begin{cases} f_1(\mathbf{x}) = (1+g(\mathbf{x}^d))(x^p + 0.02\sin(w_t\pi x^p)) \\ f_2(\mathbf{x}) = (1+g(\mathbf{x}^d))(1 - x^p + 0.02\sin(w_t\pi x^p)) \end{cases}, g(\mathbf{x}^d) = b_8(\mathbf{x}^d - G(t)) \\ G(t) = \sin(0.5\pi t), w_t = \lfloor 10G(t)\rfloor, m=2, n=50, K=1 \\ t = \dfrac{1}{n_t}\left\lfloor\dfrac{\tau}{\tau_t}\right\rfloor, n_t=10, \tau_t=20, x_i^d \in [-1,1] \end{cases}$$

40) **ETMOF40** is a 3-task optimization problem, where task $T_1$ to task $T_3$ are dynamic 2-objective optimization problem. $T_1$, $T_2$, and $T_3$ are defined as follows

$$T_1 : \begin{cases} \min \begin{cases} f_1(\mathbf{x}) = (1+g(\mathbf{x}^d))\left[x^p + 0.1\sin(3\pi x^p)\right]^{\alpha_t} \\ f_2(\mathbf{x}) = (1+g(\mathbf{x}^d))\left[1 - x^p + 0.1\sin(3\pi x^p)\right]^{\alpha_t} \end{cases}, g(\mathbf{x}^d) = b_5(\mathbf{x}^d - G(t)) \\ G(t) = \sin(0.5\pi t), \alpha_t = 0.2 + 0.28\|G(t)\| \\ m=2, n=50, K=1, t = \dfrac{1}{n_t}\left\lfloor \dfrac{\tau}{\tau_t} \right\rfloor, n_t = 10, \tau_t = 20, x_i^d \in [-1,1] \end{cases}$$

$$T_2 : \begin{cases} \min \begin{cases} f_1(\mathbf{x}) = (1+g(\mathbf{x}^d))\left[x^p + 0.1\sin(3\pi x^p)\right]^{\alpha_t} \\ f_2(\mathbf{x}) = (1+g(\mathbf{x}^d))\left[1 - x^p + 0.1\sin(3\pi x^p)\right]^{\alpha_t} \end{cases}, g(\mathbf{x}^d) = b_6(\mathbf{x}^d - G(t)) \\ G(t) = \sin(0.5\pi t), \alpha_t = 0.2 + 0.28\|G(t)\| \\ m=2, n=50, K=1, t = \dfrac{1}{n_t}\left\lfloor \dfrac{\tau}{\tau_t} \right\rfloor, n_t = 10, \tau_t = 20, x_i^d \in [-1,1] \end{cases}$$

$$T_3 : \begin{cases} \min \begin{cases} f_1(\mathbf{x}) = (1+g(\mathbf{x}^d))\left[x^p + 0.1\sin(3\pi x^p)\right]^{\alpha_t} \\ f_2(\mathbf{x}) = (1+g(\mathbf{x}^d))\left[1 - x^p + 0.1\sin(3\pi x^p)\right]^{\alpha_t} \end{cases}, g(\mathbf{x}^d) = b_9(\mathbf{x}^d - G(t)) \\ G(t) = \sin(0.5\pi t), \alpha_t = 0.2 + 0.28\|G(t)\| \\ m=2, n=50, K=1, t = \dfrac{1}{n_t}\left\lfloor \dfrac{\tau}{\tau_t} \right\rfloor, n_t = 10, \tau_t = 20, x_i^d \in [-1,1] \end{cases}$$

## 5. Experimental Settings and Performance Assessments

The following experimental settings are encouraged to use when conducting empirical studies on the proposed test suite.

1) General Settings

   **Population size**: 100 or a similar number for a single task of MOP.
   **Stopping criterion**: a maximum number of 100000 function evaluations for a single task of static MOPs, i.e., MOPs in ETMOF1 to ETMOF32, and a maximum number of 150*(31$\tau_t$) function evaluations for a single task of dynamic MOPs, i.e., the number of changes is 30 for each run of a single dynamic MOP task in ETMOF33 to ETMOF 40.
   **Number of runs:** An algorithm is required to be executed for **21** runs. Note, in each run, a new random seed should be adopted. Besides, it is prohibited to execute multiple 21 runs and then deliberately pick up the 21 better results.

2) Performance Metrics

   The IGD [17] is used to evaluate the performance of an optimizer on each task of the related static MOPs. Let $S$ be a set of non-dominated objective vectors that are obtained for task $T_k$ by the optimizer, and $S^*$ be a set of evenly distributed vectors on the true PF of task $T_k$. Then, the IGD of $S$ is computed as:

$$IGD(S, S^*) = \frac{1}{|S^*|} \sum_{x \in S^*} \min_{y \in S} d(x, y)$$

where $d(x, y)$ indicates the Manhattan distance between objective vector $x$ and $y$ in the normalized objective space. IGD can reflect the convergence and diversity of the objective vectors in $S$ simultaneously when $|S^*|$ is large enough to represent the true PF. A smaller IGD value indicates a better performance of the corresponding optimizer.

Moreover, the MIGD [18] is used to evaluate the performance of an optimizer on each task of the corresponding dynamic MOPs. Let $S_t$ be a set of non-dominated objective vectors that are obtained for task $T_k$ by the optimizer at time $t$, and $S_t^*$ be a set of evenly distributed vectors on the true PF of task $T_k$ at time $t$. The MIGD is calculated as:

$$MIGD = \frac{1}{T} \sum_{i=1}^{T} IGD(S_t, S_t^*), \; T \text{ is the number of changes in each run}$$

Finally, the mean standard score (MSS) [11] of the obtained IGD or MIGD values is used to rank the ETMO-based optimizers. Suppose there are $k$ tasks in an ETMO benchmark function, and the IGD/MIGD value obtained by an optimizer in a run for task $T_i$ is denoted as $I_i$, $i = 1, 2, \ldots, k$. In addition, suppose the average and standard deviation of the IGD/MIGD for $T_i$ are $(\mu_i, \sigma_i)$. Then the MSS of the obtained IGD/MIGD for this ETMO function is calculated as:

$$MSS = \frac{1}{k} \sum_{i=1}^{k} \frac{I_i - \mu_i}{\sigma_i}$$

In practice, $\mu_i$ and $\sigma_i$ will be computed according to the IGD/MIGD values obtained by all the optimizers on the task $T_i$ in all runs. MSS is used as a comprehensive criterion, and a smaller MSS value indicates a better overall performance of an ETMO optimizer on a benchmark function.

# References

[1] K. Deb, A. Pratap, S. Agarwal, and T. Meyarivan, "A fast and elitist multiobjective genetic algorithm: NSGA-II," *IEEE Trans. Evol. Comput.*, vol. 6, no. 2, pp. 182–197, 2002.

[2] Q.F. Zhang and H. Li, "MOEA/D: A multiobjective evolutionary algorithm based on decomposition," *IEEE Trans. Evol. Comput.*, vol. 11, no. 6, pp. 712–731, 2007.

[3] A. Gupta, Y. Ong, and L. Feng, K. C. Tan, "Multiobjective Multifactorial Optimization in Evolutionary Multitasking," *IEEE Trans. on Cybernetics*, vol. 47, no. 7, pp. 1652–1665, 2016.

[4] M. Gong, Z. Tang, H. Li, J. Zhang, "Evolutionary Multitasking with Dynamic Resource Allocating Strategy," *IEEE Trans. Evol. Comput.*, vol. 23, no. 5, pp. 858–869, 2019.

[5] J. Lin, H. Liu, K. C. Tan, and F. Gu, "An Effective Knowledge Transfer Approach for Multiobjective Multitasking Optimization", *IEEE Trans. on Cybernetics*, vol. 51, no. 6, pp. 3238-3248, 2021.

[6] A. Gupta, Y. S. Ong, L. Feng, "Insights on Transfer Optimization: Because Experience is the Best Teacher," IEEE Transactions on Emerging Topics in Computational Intelligence, vol. 2, no. 1, pp. 51-64, 2018.

[7] M. Jiang, Z. Huang, L. Qiu, W. Huang, and G. Yen, "Transfer Learning-Based Dynamic Multiobjective Optimization Algorithms", *IEEE Trans. Evol. Comput.*, vol. 22, no. 4, pp. 501–514, 2018

[8] K. C. Tan, L. Feng, M. Jiang, "Evolutionary Transfer Optimization - A New Frontier in Evolutionary Computation Research," *IEEE Computational Intelligence Magazine*, vol. 16, no. 1, pp. 22-33, 2021.
23